%% file: cvpr_arxiv.tex
\documentclass[final]{cvpr}
\input{math_commands.tex}

\usepackage{times}
\usepackage{epsfig}
\usepackage{graphicx}
\usepackage{paralist}
\usepackage{multirow}
\usepackage{floatrow}

\usepackage{graphicx}
\usepackage{amsmath}
\usepackage{amssymb}

\usepackage{xspace}

\usepackage{pifont}
\newcommand{\cmark}{\ding{51}}%
\newcommand{\xmark}{\ding{55}}%

\newcommand{\benchmark}{{NeuCRaB}\xspace}

\usepackage[pagebackref=true,breaklinks=true,colorlinks,bookmarks=false]{hyperref}

\def\cvprPaperID{7652} 
\def\confYear{CVPR 2021}

\begin{document}

\title{Ranking Neural Checkpoints}

\author{ Yandong Li $^{1,2}$\thanks{This work was done while the first author was an intern at Google.} \quad Xuhui Jia$^{2}$   \quad Ruoxin Sang$^{2}$ \\   Yukun Zhu$^{2}$  \quad Bradley Green$^{2}$  \quad Liqiang Wang$^{1}$   \quad Boqing Gong$^{2}$ \\
$^{1}$University of Central Florida  \quad $^{2}$Google\\
{\tt\small lyndon.leeseu@outlook.com \quad lwang@cs.ucf.edu \quad \{xhjia,rxsang,yukun,brg,bgong\}@google.com  }\\
}

\maketitle

\begin{abstract}
This paper is concerned with ranking many pre-trained deep neural networks (DNNs), called checkpoints, for the transfer learning to a downstream task. Thanks to the broad use of DNNs, we may easily collect hundreds of checkpoints from various sources. Which of them transfers the best to our downstream task of interest? Striving to answer this question thoroughly, we establish a neural checkpoint ranking benchmark (\benchmark) and study some intuitive ranking measures. These measures are generic, applying to the checkpoints of different output types without knowing how the checkpoints are pre-trained on which datasets. They also incur low computation cost, being practically meaningful. Our results suggest that the linear separability of the features extracted by the checkpoints is a strong indicator of transferability. We also arrive at a new ranking measure, $\mathcal{N}$LEEP, which gives rise to the best performance in the experiments. 
Code is available at \url{https://github.com/google-research/google-research/tree/master/rank_ckpt}.
\end{abstract}

\section{Introduction}
There is an increasing number of pre-trained deep neural networks (DNNs), which we call checkpoints. We may produce hundreds of intermediate checkpoints when we sweep through various learning rates, optimizers, and losses to train a DNN. Furthermore,  semi-supervised~\cite{chapelle2009semi,bachman2014learning,rasmus2015semi,laine2016temporal,tarvainen2017mean,miyato2018virtual,luo2018smooth,berthelot2019mixmatch} and self-supervised~\cite{doersch2015unsupervised,he2020momentum,chen2020simple,veeling2018rotation,noroozi2016unsupervised} learning make it feasible to harvest DNN checkpoints with scarce or no labels. Fine-tuning~\cite{yosinski2014transferable, pan2009survey} has become a de facto standard to adapt the pre-trained checkpoints to target tasks. It leads to faster convergence~\cite{donahue2014decaf, he2019rethinking, sharif2014cnn} and better performance~\cite{kornblith2019better} on the downstream tasks. 

However, not all checkpoints are equally useful for a target task, and some could even under-perform a randomly initialized checkpoint (cf.\ Section~\ref{sec:checkpoints}). This paper is concerned with \textbf{ranking neural checkpoints}, which aims to measure how effectively fine-tuning can transfer knowledge from the pre-trained checkpoints to the target task. The measurement should be \emph{generic} enough for all the neural checkpoints, meaning that it works without knowing any pre-training details (e.g., pre-training examples, hyper-parameters, losses, early stopping stages, etc.) of the  checkpoints. It also should be \emph{lightweight}, ideally without training on the downstream task, to make it practically useful. We may use the measurement to choose the top few checkpoints before running fine-tuning, which is computationally more expensive than calculating the measurements.

Ranking neural checkpoints is crucial. 
Some domains or applications lack large-scale human-curated data, like medical images~\cite{raghu2019transfusion}, raising a pressing need for high-quality pre-trained checkpoints as a warm start for fine-tuning. Fortunately, there exist hundreds of thousands of checkpoints of popular neural network architectures. For instance, many computer vision models are built upon ResNet~\cite{he2016deep}, Inception-ResNet~\cite{szegedy2016inception}, and VGG~\cite{simonyan2014very}. As a result, we can construct a candidate pool by collecting the checkpoints released by different groups, for various tasks, and over distinct datasets.

It is nontrivial to rank the checkpoints for a downstream task. We explain this point by drawing insights from the related, yet arguably easier, task transferability problem~\cite{task2vec,eaton2008modeling,zamir2018taskonomy,leep}, which aims to provide high-level guidance about how well a neural network pre-trained in one task might transfer to another. However, not all checkpoints pre-trained in the same source task transfer equally well to the target task~\cite{zoph2020rethinking, kornblith2019better}. The pre-training strategy also matters. Zhai~\textit{et al.}~\cite{zhai2019visual} find that combining supervision with self-supervision improves a network's transfer results on downstream tasks. He~\textit{et al.}~\cite{he2020momentum} also show that self-supervised pre-training benefits object detection more than its supervised counterpart under the same fine-tuning setup. 


We may also appreciate the challenge in ranking neural checkpoints by comparing it with another related line of work: predicting DNNs' generalization gaps~\cite{neyshabur2017exploring,kawaguchi2017generalization,bartlett2017spectrally}. Jiang~\textit{et al.}~\cite{jiang2018predicting} use a linear regressor to predict a DNN's generalization gap, i.e., the  discrepancy between its training and test accuracies, by exploring the training data's margin distributions. Other signals studied in the literature include network complexity and noise stability. Ranking neural checkpoints is more challenging than predicting a DNN's generalization gap. Unlike the training and test sets that share the same underlying distribution, the downstream task may be arbitrarily distant from the source task over which a checkpoint is pre-trained. Moreover, we do not have access to the pre-training data at all. Finally, instead of keeping the networks static, fine-tuning dramatically changes all weights of the checkpoints.



We establish a neural checkpoint ranking benchmark (\benchmark) to study the problem systematically. \benchmark covers various checkpoints pre-trained on widely used, large-scale datasets by different training strategies and architectures at a range of early stopping stages. It also contains diverse downstream tasks, whose training sets are medium-sized, making it practically meaningful to rank and fine-tune existing checkpoints. Pairing up all the checkpoints and downstream tasks, we conduct careful fine-tuning with thorough hyper-parameter sweeping to obtain the best transfer accuracy for each checkpoint-downstream-task pair. Hence, we know the groundtruth ranking of the checkpoints for each downstream task according to the final accuracies (over the test/validation sets). 

A functional checkpoint ranking measurement should be highly correlated with the groundtruth ranking and, equally importantly, incurs as low computation cost as possible. We study several intuitive methods for ranking the neural checkpoints. One is to freeze the checkpoints as feature extractors and use a linear classifier to evaluate the features' separability on the target task. Another is to run fine-tuning for only a few epochs (to avoid heavy computation) and then evaluate the resulting networks on the target task's validation set. We also estimate the mutual information between labels and the features extracted from a checkpoint.



Finally, we propose a lightweight measure, named Gaussian LEEP ($\mathcal{N}$LEEP), to rank checkpoints based on the recently proposed log expected empirical prediction (LEEP)~\cite{leep}. LEEP was originally designed to measure between-task transferabilities. It cannot handle the checkpoints pre-trained by unsupervised or self-supervised learning since it requires all checkpoints to have a classification head. Its computation cost could blow up when the classification head corresponds to a large output space. Moreover, it depends on the classification head's probabilistic output, which, unfortunately, is often overly confident~\cite{guo2017calibration}.


To tackle the above problems, we replace the checkpoints' output layer with a Gaussian mixture model (GMM). This simple change kills two birds with one stone. On the one hand, GMM's soft assignment of  input to clusters seamlessly applies to LEEP, resulting in the lightweight, effective $\mathcal{N}$LEEP measure that works regardless of the checkpoints' output types. On the other hand, since we fit GMM to the target task's data, instead of the pre-training data of a different source task, the cluster assignment probabilities are likely more calibrated than the classification probabilities for the target task, if there exist classification heads.

\section{The Neural Checkpoint Ranking Benchmark (\benchmark)}\label{sec:ncrb}
We formalize ranking neural checkpoints as follows. Suppose we have $m$ pre-trained neural networks, called checkpoints, $\mathcal{C}:=\{\theta_i\}_{i=1}^m$. Denote by $\mathcal{T}$ a distribution of tasks. Without loss of generality, we mainly study classification downstream tasks, each of which, $t\sim \mathcal{T}$, contains a training set and a test set. 
An evaluation procedure, $\mathbf{G}:\mathcal{C}\times\mathcal{T}\mapsto\mathbb{R}$, replaces the output layer of a checkpoint $\theta_i$ with a linear classifier for a downstream task $t$, followed by fine-tuning using the task's training set. It employs hyper-parameter sweeping and returns the best accuracy on the test set. We apply this evaluation procedure to all the checkpoints for  task $t$ and obtain their test accuracies:
\begin{align}
\mathbf{G}_t := \{\mathbf{G}(\theta_i, t)\}_{i=1}^m \in \mathbb{R}^m, \label{eq:eval}
\end{align}
which defines the groundtruth ranking list for task $t$.

Denote by $\mathcal{R}$ all measures that return a ranking score for any checkpoint-task pair under a computation budget $\mathbf{b}$. A measure $\mathbf{R}\in \mathcal{R}$ gives rise to the following ranking scores for a task $t$,
\begin{align}
\mathbf{R}_t:=\{\mathbf{R}(\theta_i,t;\mathbf{b})\}_{i=1}^m \in \mathbb{R}^m,
\end{align}
where we underscore the computation budget $\mathbf{b}$ in the measure $\mathbf{R}(\cdot,\cdot;\mathbf{b})$. 

\begin{figure*}
\centering
\includegraphics[width = 1\textwidth]{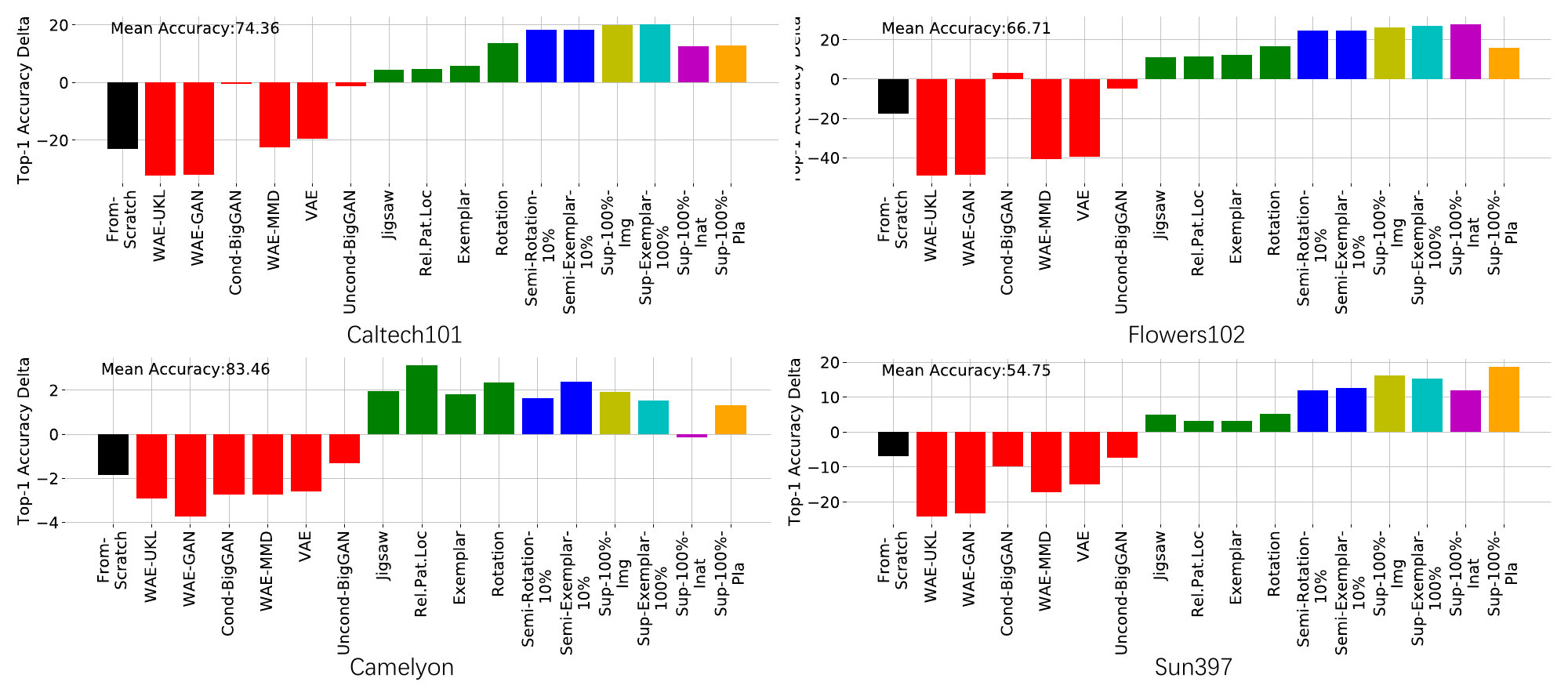}
\caption{Fine-tuning the checkpoints in Group I on four downstream tasks. We keep the best fine-tuning accuracy for each checkpoint-task pair after hyper-parameter sweeping. For better visualization, the values are offset by their mean (cf.\ Table 4 in Appendix for the absolute values). (Best viewed in color. Red: generative models. Black: From-Scratch. Green: self-supervised models. Blue: semi-supervised models. Yellow, Pink, and Orange: supervised models trained on ImageNet, Inatualist, and Places365, respectively. Cyan: a hybridly-supervised model.) }
\label{fig:groundtruth}
\end{figure*}
Our objective in ranking neural checkpoints is to find the best ranking measure in expectation,
\begin{align} \label{eq:objective}
\mathbf{R}^* \leftarrow \argmax\limits_{\mathbf{R} \in \mathcal{R}}\;\mathbb{E}_{t \sim \mathcal{T}}\; \mathcal{M}(\mathbf{R}_t, \mathbf{G}_t)
\end{align}
where $\mathcal{M}$ is a metric evaluating the ranking scores $\mathbf{R}_t$ against the test accuracies $\mathbf{G}_t$. Section~\ref{sec-eval} details the evaluation methods used in this work. Equipped with such a ranking measure $\mathbf{R}^*$, we can identify the checkpoints that potentially transfer to a downstream task better than the others without resorting to heavy computation. 

\subsection{Downstream Tasks $\mathcal{T}$} \label{sec:tasks}
Following the design principle of~\cite{zhai2019visual}, we study diverse downstream tasks including Caltech101~\cite{fei2006one}, Flowers102~\cite{nilsback2008automated},  Sun397~\cite{xiao2010sun}, and Patch Camelyon~\cite{veeling2018rotation}. These tasks are representative of general object recognition, fine-grained object recognition, scenery image classification, and medical image classification, respectively. Table 1 in Appendix A.1 provides more details of these tasks. A common theme is that their training sets are all medium-sized, making it especially beneficial  to leverage pre-trained checkpoints to avoid overfitting.



\subsection{Neural Checkpoints $\mathcal{C}$} \label{sec:checkpoints}
Thanks to the broad use of DNNs, one may collect neural checkpoints of various types from multiple sources. To simulate this situation, we construct a rich set of checkpoints and  separate them into three groups according to the pre-training strategies and network architectures.

\textbf{Group I: Checkpoints of mixed supervision.}
The first group of checkpoints are pre-trained with mixed supervision till convergence, including supervised learning, self-supervised learning, semi-supervised learning, and the discriminators or encoders in deep generative models. It consists of 16 ResNet-50s~\cite{he2016deep}. We borrow 14 models pre-trained on ImageNet~\cite{deng2009imagenet} from~\cite{zhai2019visual}. Among them, four are pre-trained by self-supervised learning (Jigsaw~\cite{noroozi2016unsupervised}, Relative Patch Location~\cite{doersch2015unsupervised}, Exemplar~\cite{dosovitskiy2014discriminative}, and Rotation~\cite{gidaris2018unsupervised}), six are the discriminators of generative models (WAE-UKL~\cite{rubenstein2019practical}, WAE-GAN, WAE-MMD~\cite{tolstikhin2017wasserstein}, Cond-BigGAN, Uncond-BigGAN~\cite{brock2018large}, and VAE~\cite{kingma2013auto}), two are based on semi-supervised learning (Semi-Rotation-10\% and Semi-Exemplar-10\%~\cite{zhai2019s4l}), one is by fully supervised learning (Sup-100\%-Img~\cite{he2016deep}), and one is trained with a hybrid supervised loss (Sup-Exemplar-100\%~\cite{zhai2019s4l}). We also add two supervised checkpoints pre-trained on iNaturalist (Sup-100\%-Inat)~\cite{van2018inaturalist} and Places365 (Sup-100\%-Pla)~\cite{zhou2017places}, respectively. Using the evaluation procedure $\mathbf{G}(\theta_i,t)$ (cf.\ equation~(\ref{eq:eval})), we obtain their final accuracies on the downstream tasks described in Section~\ref{sec:tasks}.

Figure~\ref{fig:groundtruth} shows the best fine-tuning accuracies offset by their mean for better visualization, and Table 4 (in Appendix) contains the absolute accuracy values. We include the training from scratch (From-Scratch) for comparison. Most of the checkpoints yield significantly better fine-tuning results than From-Scratch. Some of the discriminators in generative models, however, under-perform From-Scratch. The highest-performance checkpoints change from one downstream task to another. 

\textbf{Group II: Checkpoints at different pre-training stages.}
This group comprises 12 ResNet-50s pre-trained by fully supervised learning on ImageNet, iNaturalist, and Places-365. We save a checkpoint right after each learning rate decay, resulting in four checkpoints per dataset. Figure 2 and Table 5 in Appendix show the best fine-tuning accuracies over the four downstream tasks, where Img-90k refers to the checkpoint trained on ImageNet for 90k iterations. Interestingly, the downstream tasks favor different pre-training sources, indicating the necessity of studying between-task transferabilities~\cite{zhai2019visual,zamir2018taskonomy}. However, the source task information may be not known for all checkpoints. Moreover, the converged model over a source task does not always transfer the best to a downstream task (cf.\ Img-270k vs.\ Img-300k on Camelyon, Inat-270k vs.\ Inet-300k on Flowers102, etc.). We hence construct this \benchmark for studying the ranking of neural checkpoints without accessing how one pre-trained the checkpoints over which dataset.



\textbf{Group III: Checkpoints of heterogeneous architectures.} Kornblith~\textit{et al.}~\cite{kornblith2019better} show that better network architectures can learn better features that can be transferred across vision-based tasks.
Therefore, we construct the third group of checkpoints by using different neural architectures. Four of them belong to the Inception family~\cite{szegedy2015going}, one is Inception-ResNet-v2~\cite{szegedy2016inception}, six come from the MobileNet family~\cite{howard2017mobilenets}, and two are from the ResNet-v1 family~\cite{he2016deep}. We train them on ImageNet till convergence. Figure 3 and Table 6 in Appendix visualize their fine-tuning accuracies on the four downstream tasks. 
\subsection{Evaluation Metrics $\mathcal{M}$} \label{sec-eval}
We use multiple metrics  (cf.\ $\mathcal{M}$ in eq.~(\ref{eq:objective})) to evaluate the checkpoint ranking measures. 


\begin{description}
\setlength\itemsep{-0pt}
\item[Recall@$k$:] A practitioner may have resources to test up to $k$ checkpoints for their task of interest. We consider it a success if a measure ranks the highest-performance checkpoint into the top $k$. A measure's Recall@$k$ is the ratio between the number of downstream tasks on which it succeeds and the total number of tasks. We employ $k=1$ and $k=3$ in the experiments.



\item[Top-$k$ relative accuracy (Rel@$k$):] Given a task, a ranking measure returns an ordered list of the checkpoints. If the measure selects a high-performing checkpoint to the top $k$ despite that it misses the highest-performance one, we do not want to overly penalize it. 
This Rel@$k$ is the ratio between the best fine-tuning accuracy on the downstream task with the top $k$ checkpoints and the the best fine-tuning accuracy with all the checkpoints.



\item[Pearson correlation:]  We incorporate Pearson's $r$~\cite{pearson1895vii} to compute the linear correlation between a measure' ranking scores $\mathbf{R}_t$ and the evaluation procedure's final accuracies $\mathbf{G}_t$. 

\item[Kendall ranking correlation:] We also include  Kendall's $\tau$~\cite{kendall1938new} to measure the ordinal association between a ranking measure $\mathbf{R}$ and the evaluation procedure $\mathbf{G}$ for each task. After all, what matter is the order of the checkpoints rather than the precise ranking scores.


\end{description}
\section{Checkpoint Ranking Methods}
In this section, we describe some intuitive neural checkpoint ranking methods. These methods strive to achieve high correlation with the checkpoint evaluation procedure $\mathbf{G}$ at low computation cost.  



\subsection{Fine-tuning with Early Stopping}
If there is no constraint over computing, the evaluation procedure $\mathbf{G}$ itself becomes the gold ranking measure. Hence, a natural ranking method is the fine-tuning with early stopping, by which the model is far from  convergence. The premature models' test accuracies are the ranking scores. Experiments reveal that it is hard to forecast from the premature models. 

\subsection{Linear Classifiers}
We derive the second ranking method also from the evaluation procedure $\mathbf{G}$, which replaces a checkpoint's output layer by a linear  classifier tailored for the downstream task. We train the linear classifier while freezing the other layers. The ranking score equals the classifier's test accuracy. It is worth mentioning that self-supervised learning~\cite{chen2020simple,he2020momentum,grill2020bootstrap} often adopts this practice as well to evaluate the learned feature representations. We shall see that the linear separability of the features extracted from a checkpoint is a strong indicator of the performance of fine-tuning the full checkpoint.


\subsection{Mutual Information}
Suppose the extracted features' quality well correlates with a checkpoint's final accuracy on a downstream task. Besides the linear separability above, we can rank the checkpoints by their mutual information between the high-dimensional features and discrete labels of the downstream task. We employ the state-of-the-art $I_\alpha$ mutual information estimator~\cite{poole2019variational}, where $\alpha$ controls the trade-off between variance and bias. It is a variational lower bound parameterized by a neural network. Belghazi~\textit{et al.}~\cite{belghazi2018mine} report that the neural estimators generally outperform prior mutual information estimations, especially when the  variables are high-dimensional. We use the code released by the authors to calculate $I_\alpha$~\cite{poole2019variational}.

\subsection{LEEP for the Checkpoints with Classification Heads}
To rank the checkpoints pre-trained over classification source tasks, the recently proposed LEEP~\cite{leep} measure is directly applicable despite that it was originally designed for between-task transfer. Denote by $\mathcal{Z}$ the classification space of a checkpoint $\theta$. We can interpret $\theta(x)_z$,  the $z$-th (softmax) output element, as the probability of classifying the input $x$ into the class $z\in\mathcal{Z}$. Given a downstream task $t\sim \mathcal{T}$ and its test set $\{(x_j,y_j)\}_{j=1}^n$, the LEEP ranking score for the checkpoint $\theta$ is calculated by
\begin{align}
\nonumber
\mathbf{R}_\textsc{LEEP}(\theta,t) &:= \frac{1}{n}\sum_{j=1}^n \log P(y_j|x_j,\theta,t) \\ P(y|x,\theta,t) &:=\sum_{z\in\mathcal{Z}}\hat{P}(y|z)\theta(x)_z \label{eq:leep}
\end{align}
where $\hat{P}(y|z)$ is the empirical conditional distribution of the downstream task's label $y$ given the source label $z\in\mathcal{Z}$, and $P(y|x,\theta,t)$ is a ``dummy'' classifier, which firstly draws a label $z$ from the checkpoint $\theta(x)$ and then draws a class $y$ from the conditional distribution $\hat{P}(y|z)$. 

Denote by $\{x_j,y_j\}_{j=1}^{\Tilde{n}}$, $y\in\mathcal{Y}$, the downstream task's training set. LEEP computes the  conditional distribution $\hat{P}(y|z)$ by ``counting''. The joint distribution $\hat{P}(y,z)$ due to the checkpoint $\theta$ is
\begin{align}
\hat{P}(y, z) = \frac{1}{\Tilde{n}}\sum_{j:y_j=y} \theta(x_j)_z, \label{eq:joint}
\end{align}
which gives rise to the conditional distribution $\hat{P}(y|z) = {\hat{P}(y,z)}/{\hat{P}(z)} =  {\hat{P}(y,z)}/{\sum_{y\in\mathcal{Y}} \hat{P}(y,z)}$. 



In the experiments, LEEP and the linear classifier are the second best ranking methods for the checkpoints pre-trained for classification. However, LEEP's computation cost is high when a checkpoint's classification output is high-dimensional (e.g., iNaturalist contains more than 8000 classes). Besides, its softmax estimation of the classification probability $\theta(x)_z$ is often poorly calibrated~\cite{guo2017calibration}. Finally, it does not apply to the checkpoints with no classification heads.


\subsection{$\mathcal{N}$LEEP}
We propose a variation to LEEP that applies to all types of checkpoints including those obtained from unsupervised learning and self-supervised learning. It can also avoids the overly confident softmax. 

Feeding the training data of a downstream task into a checkpoint, we obtain their feature representations. The representations are thousands of dimensions, depending on the checkpoint's neural architecture. We reduce their dimension by using the principal component analysis (PCA). Denote by $s$ the resultant low-dimensional representation of the input $x$. 

We then fit a Gaussian mixture model (GMM), $P(s) = \sum_{v\in \mathcal{V}} \pi_v \mathcal{N}(s| \mu_v, \Sigma_v)$, to the training set $\{s_j\}_{j=1}^{\Tilde{n}}$, where $\mathcal{V}$ is a collection of all the Gaussian components, and $\pi_v,v\in\mathcal{V},$ are the mixture weights. It is convenient to compute the posterior distribution: 
\begin{align}
    P(v|x)=P(v|s) \propto  \pi_v \mathcal{N}(s| \mu_v, \Sigma_v),
\end{align}
which is arguably more reliable than the class assignment probability $\theta(x)_z$ output by the softmax classifier because we fit GMM to the downstream task's training data, whereas the softmax classifier is learned from a different source task. 

Hence, we arrive at an improved ranking measure, named $\mathcal{N}$LEEP, by replacing  $\theta(x)_z$, the probability of classifying an input $x$ to the class $z$, in equations~(\ref{eq:leep}--\ref{eq:joint}) by the posterior distribution $P(v|x)$.


\begin{figure*}
\centering

\includegraphics[width = 1.0\textwidth]{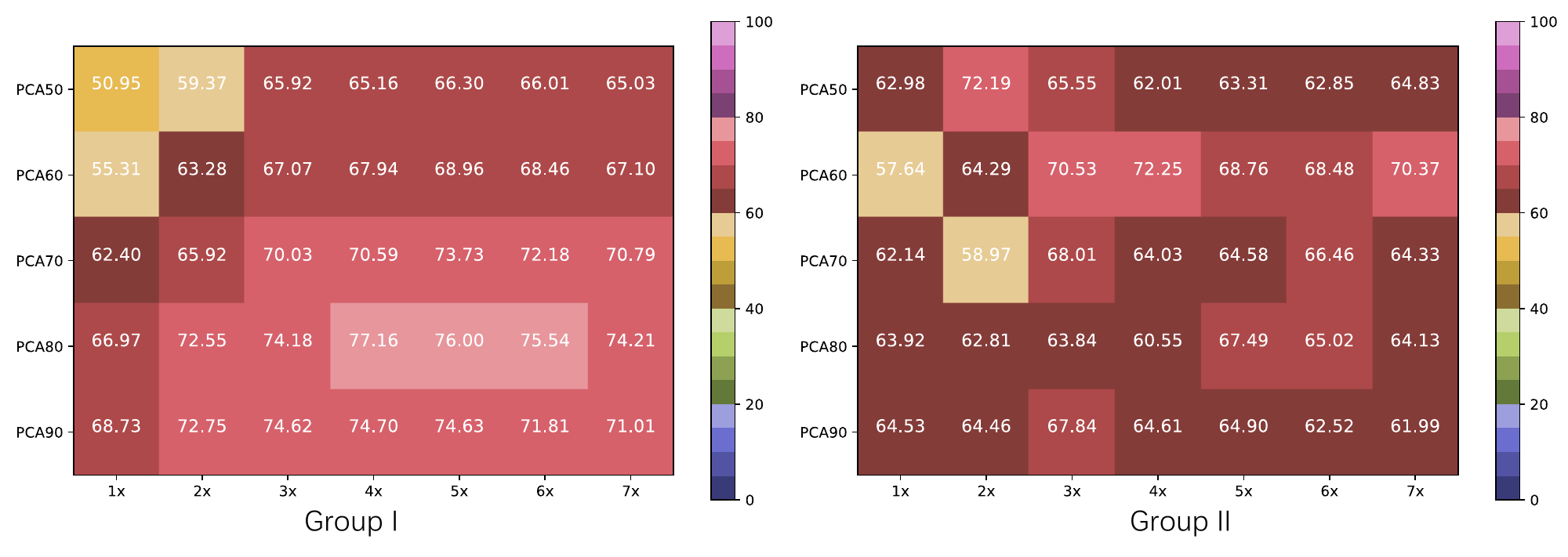}
\vspace{-10pt}
\caption{$\mathcal{N}$LEEP' checkpoint ranking performance, evaluated by Kendall's $\tau$, on Groups I and II in \benchmark. We vary the PCA feature dimension and the number of Gaussian components in GMM.}
\label{fig:mixedandsup_overall}
\vspace{-5pt}
\end{figure*}

\section{Experiments on \benchmark}\label{sec:exp}
There are free parameters in each of the  ranking methods. Before presenting the main  results, we study how the free parameters in $\mathcal{N}$LEEP affect its checkpoint ranking performance. Figure~\ref{fig:mixedandsup_overall} illustrates $\mathcal{N}$LEEP's Kendall’s $\tau$ values over {Groups I} and {II} with different PCA feature dimensions and the numbers of Gaussian components. Each Kendall's $\tau$ is averaged across all the downstream tasks; the higher, the better. Along the vertical axes, we change the feature dimensions by keeping different percentages of the PCA energies;  PCA50 means the percentage is 50\%. Along the horizontal axes, we adopt different numbers of Gaussian components in GMM; $2\times$ means the number is twice the class number of the downstream task. Notably, the Kendall's $\tau$ values remain relatively stable. In the remaining experiments with $\mathcal{N}$LEEP, we fix the PCA energy to 80\% and the number of Gaussian components five times the class number of a downstream task. 


\subsection{Comparison Results}
Tables~\ref{tab:comparision_mixedsup},~\ref{tab:comparision_sup},~and~\ref{tab:comparision_arch} show the checkpoint ranking methods' performance on Groups I (checkpoints of mixed supervision), II (different pre-training stages), and III (heterogeneous architectures), respectively. We also union the three groups and present the corresponding ranking performance in Table 2 in Appendix. The numbers in the tables are the average over all downstream tasks. In addition to the evaluation metrics detailed in Section~\ref{sec-eval}, the GFLOPS column  measures the ranking methods' computing performance; the lower, the better.

We report multiple variations of the ranking methods in the tables. Fine-tuning is computationally expensive, so we stop it after one or five epochs. The linear classifiers are less so as we save the feature representations of downstream tasks' after one forward pass to the checkpoints. We report the linear classifiers' ranking results after training them for one epoch,  five epochs, and convergence. We test $\alpha=0.01$ and $\alpha=0.50$ in the $I_\alpha$ mutual information estimator. Additionally, we experiment with $I_\alpha$ after reducing the feature dimensions by using PCA. 


\subsection{Main Findings} In each column of Tables~\ref{tab:comparision_mixedsup},~\ref{tab:comparision_sup},~\ref{tab:comparision_arch},~and Table 2 in Appendix, we highlight the best and second best by the bold font and underscore, respectively. 

\vspace{3pt}
\noindent\textbf{The mutual information} fails to rank high-performing checkpoints to the top and even produces negative Pearson and Kendall correlations, probably because of the features' high dimensions. Reducing the feature dimensions by PCA significantly improves the mutual information's ranking performance; MI w/ PCA ($\alpha$=0.01) leads to the second best Rel@1, Recall@3 and Rel@3 among the ranking methods in Group III, the checkpoints of heterogeneous neural architectures.  Varying $\alpha$ in the $I_\alpha$ mutual information estimator~\cite{poole2019variational} can control the trade-off between variance and bias. MI w/ and w/o PCA ($\alpha$=0.01) perform better than MI w/ and w/o PCA ($\alpha$=0.50), respectively. It indicates that neural checkpoint ranking requires low-bias MI estimator since smaller $\alpha$ means low-bias but high-variance estimation.

\vspace{3pt}
\noindent\textbf{Fine-tuning} up to some epochs turns out the worst ranking methods because it leads to low correlation with the groundtruth ranking and yet incurs heavy computation. Similarly, training the linear classifier up to one or five epochs does not perform well except in Group II.   These results indicate that it is difficult to forecast the checkpoints' final performance from  premature models. Fine-tuning (5 epochs) and Linear (5 epochs) perform better than Fine-tuning (1 epoch) and Linear (1 epoch) in terms of Person and Kendall correlation, respectively. However, they all fail to select the top checkpoint in Group I and Group III since they produce lower Recall@1 and Recall@3 than others. One possible reason is that the evaluation accuracies of checkpoints in the early stage tend to have large variance.

\vspace{3pt}
\noindent\textbf{Feature qualities before fine-tuning the checkpoints.}
If we train the linear classifiers till convergence, they become the best in Group II, and the second best checkpoint ranking method in Groups I and III in terms of Pearson and Kendall correlations. It can also produce better Recall@1 and Recall@3 than Linear (1 epoch) and Linear (5 epoch) in Groups I, II and III since the evaluation accuracies of converged models are more stable than models in the early training stage. Note that the linear classifiers' accuracies, i.e., the ranking scores, imply the linear separability of the features extracted by the checkpoints. Recall that the mutual information with PCA feature dimension reduction is among the second best (Rel@1, Recall@3 and Rel@3) in Group III. Since both methods measure the feature representations' quality by the downstream tasks' labels, we conjecture that the quality of the features is a strong indicator of the checkpoints' final fine-tuning performance on the downstream tasks. It would be interesting to study other feature quality measures beyond the linear separability and mutual information in future work. 
\input{tables-three-groups}

\vspace{3pt}
\noindent\textbf{
$\mathcal{N}$LEEP} performs consistently well in all the groups of checkpoints over all the evaluation metrics with the lowest computation cost . In contrast, the original LEEP measure is not applicable to Group I, the checkpoints of mixed supervision, because it requires that the checkpoints have a classification output layer. Overall, LEEP is the second best over all evaluation metrics among the ranking methods in Groups II and III, whose checkpoints all have a classification output layer. Specifically, LEEP can produce the second best Recall@1, Recall@3 and Rel@3 in Group II, and the best Recall@3, the best Rel@3 and the second best Kendall correlation in Group III. It is a more consistent indicator than fine-tuning, linear classifier, or MI based ranking methods. However, LEEP can not produce better results than $\mathcal{N}$LEEP, and it requires slightly larger GFLOPS due to the extra computation cost from the classification head. 

We conjecture that $\mathcal{N}$LEEP outperforms LEEP mainly because GMMs calibrate the posterior probabilities better than the checkpoints' softmax classifiers. The checkpoint ranking quality of LEEP score hinges on the performance of the `dummy classifier' -- $P(y|x,\theta,t)$, and $\theta(x)_z$ is the key element to calculate it. However, $\theta(x)$ can be poorly calibrated~\cite{leep} and it can not represent a true probability. In contrast, $P(v|x)$ used in $\mathcal{N}$LEEP is indeed the probability that the sample belongs to one cluster from a mixture of Gaussian distributions and it can remedy the poor-calibrated problem in LEEP. 

\vspace{3pt}
\noindent\textbf{Computational costs.}
Moreover, we highlight the GFLOPS column in the tables. $\mathcal{N}$LEEP and LEEP exhibit a clear advantage over the other checkpoint ranking methods in terms of computing. The main reason is that $\mathcal{N}$LEEP and LEEP can avoid intensive computation from neural network training, and they only require one forward pass through the training data.

\vspace{3pt}
\noindent\textbf{Comparing different groups of the checkpoints.}
Checkpoint ranking on different groups of checkpoints varies in degrees of difficulty. The most challenging group is Group III, the checkpoints of heterogeneous neural architectures. All the ranking methods produce lower correlations with the groundtruth ranking, and they can barely select the top checkpoints in this group. The main reason is that the neural architectures matter for transfer learning~\cite{kornblith2019better}. Besides, heterogeneous neural architectures can demonstrate various performance even if we train them from scratch on downstream tasks. Ranking neural checkpoints by the feature representations of the last layer is not sufficient for those checkpoints. We may explore more advanced ranking methods considering the structures of the deep neural networks in the future.

Checkpoint ranking on Group II is easier than on Group I since all the ranking methods can achieve relatively better results over all evaluation metrics in Group II. The results indicate that checkpoints with various training strategies (Group I) can bring more complex knowledge from source domains,  comparing with checkpoints with different early stopping stages (Group II). In addition, fine-tuning the entire models and training linear classifiers up to one or five epochs perform significantly better on Group II since those ranking methods are based on early stopping as well. 

\vspace{3pt}
{\noindent \bf Additional experiments in the supplementary materials.} 
To simulate a sufficiently large pool of checkpoints in the real applications, we finally combine the checkpoints in Group I, II, and III into one large group and conduct checkpoint ranking experiments on it. We also add one more group of checkpoints with ResNet-101s~\cite{he2016deep} to evaluate the checkpoint ranking on deeper models. Please see more details in Appendix A.3 and A.4. We also take object detection and instance segmentation as downstream tasks and conduct preliminary experiments on VOC~\cite{everingham2010pascal} and Cityscapes~\cite{Cordts2016Cityscapes}. Please refer to Appendix A.6 to see detailed discussions.

\vspace{3pt}
Although the benchmark can be easily extended to many downstream tasks in other modalities, e.g., voice, text, and cross-modal modalities, we steer our attention into comparing several intuitive ranking measures on the variants of checkpoints, covering different training strategies, source domains, and architectures at a range of early stopping stages. We formalize the checkpoint ranking idea, demonstrate the existence of an effective yet lightweight measure, $\mathcal{N}$LEEP, and hope it can shed light on more efficient ranking methods and practical applications.

\input{related}

















\section{Conclusion}
Deep learning has triumphed over many fields in both research and real-world applications. There must exist hundreds of thousands of DNNs trained and released by various groups. To this end, it is natural to select an existing, promising DNN checkpoint as a warm start to a training procedure when solving a new task. How to identify useful checkpoints from a large pool for the target task? Towards answering this question, we present \benchmark, a thorough benchmark covering diverse downstream tasks and pre-trained DNN checkpoints, along with $\mathcal{N}$LEEP, a lightweight, effective checkpoint ranking measure. 

The experiments with linear classifiers and  mutual information (after PCA) reveal that the features extracted from the checkpoints are good indicators of the checkpoints' potential in transfer learning. It is worth exploring other ways of evaluating the features' quality in future work. It is also interesting to investigate the checkpoints' inherent signatures, such as topology and stability to noise, which might be informative of their transferabilities. Finally, some learning-based methods in predicting networks' generalization gaps are also promising for the checkpoint ranking problem. 



{\small
\bibliographystyle{ieee_fullname}

}

\appendix

\newpage

\input{supp_concat}

\end{document}

%% file: math_commands.tex
\usepackage{amsmath,amsfonts,bm}

\def\eqref#1{equation~\ref{#1}}

\def\1{\bm{1}}

\DeclareMathAlphabet{\mathsfit}{\encodingdefault}{\sfdefault}{m}{sl}
\SetMathAlphabet{\mathsfit}{bold}{\encodingdefault}{\sfdefault}{bx}{n}

\DeclareMathOperator*{\argmax}{arg\,max}

%% file: tables-three-groups.tex
\begin{table*}[h!]
\centering
\caption{Checkpoint ranking results on Group I, the checkpoints of mixed supervision (GFLOPS excludes a forward pass on training data, which takes 3.04E5 GFLOPS shared by all methods)}
\label{tab:comparision_mixedsup}

\begin{tabular}{l|c|c|c|c|c|c|c}
\hline
 Method & Recall@$1$  & Rel@$1$  & Recall@$3$ & Rel@$3$  & Pearson & Kendall & GFLOPS  \\
 
  
\hline
   Linear (1 epoch)& 0.00 & 96.97 & 25.00 & 98.79 & 23.56 & 18.44 &  \underline{4.95E4}\\
   Linear (5 epoch)& 25.00 & 98.79 & \underline{50.00} & 98.94 & 49.77 & 32.33 &  4.97E4\\

 Linear (converged) & \underline{50.00} & \underline{99.63} & \textbf{75.00} & \textbf{99.65}	& \underline{68.97}	& \underline{53.43} & 5.33E4 \\
 
  Fine-tune (1 epoch)& 25.00 & 97.45 & 25.00 & 97.66 & 30.25 & 22.15 &6.51E5 \\
  Fine-tune (5 epoch)& 0.00 & 91.09 & 25.00 & 98.61 & 48.19 & 36.78 & 4.28E6\\
\hline  

MI ($\alpha$=0.01)~\cite{poole2019variational} & \multirow{1}{*}{0.00} & \multirow{1}{*}{64.67} & \multirow{1}{*}{0.00} & \multirow{1}{*}{87.96} & \multirow{1}{*}{2.39} & \multirow{1}{*}{-0.31} &\multirow{1}{*}{1.62E5} \\
 
MI ($\alpha$=0.50)  & \multirow{1}{*}{0.00} & \multirow{1}{*}{66.71} & \multirow{1}{*}{25.00} & \multirow{1}{*}{90.31} &\multirow{1}{*}{-4.91}  & \multirow{1}{*}{-13.05} &1.62E5\\

MI w/ PCA ($\alpha$=0.01) & \multirow{1}{*}{0.00} & \multirow{1}{*}{89.45} & \multirow{1}{*}{50.00} & \multirow{1}{*}{\underline{99.27}} & \multirow{1}{*}{16.16} & \multirow{1}{*}{20.67} &5.58E4\\

MI w/ PCA ($\alpha$=0.50) & \multirow{1}{*}{0.00} & \multirow{1}{*}{86.49} & \multirow{1}{*}{25.00} & \multirow{1}{*}{94.28} & \multirow{1}{*}{-24.72} & \multirow{1}{*}{-16.06} &5.58E4\\  
\hline
LEEP~\cite{leep} & \multirow{1}{*}{--} & \multirow{1}{*}{--} & \multirow{1}{*}{{--}} & \multirow{1}{*}{{--}} & \multirow{1}{*}{--} & \multirow{1}{*}{--} & \multirow{1}{*}{--}\\ 
\hline
$\mathcal{N}$LEEP & \textbf{75.00} & \textbf{99.65} & \textbf{75.00} & \textbf{99.65} & \textbf{84.30} & \textbf{76.00} & \textbf{12.85}\\ 

\hline
\end{tabular}
\end{table*}

\begin{table*}[h]
\centering
\caption{Checkpoint ranking results on Group II, the  checkpoints at different pre-training stages (GFLOPS excludes a forward pass on training data, which takes 3.04E5 GFLOPS shared by all)}
\label{tab:comparision_sup}

\begin{tabular}{l|c|c|c|c|c|c|c}
\hline
  Method & Recall@$1$  & Rel@$1$  & Recall@$3$ & Rel@$3$  & Pearson & Kendall & GFLOPS  \\
\hline
  Linear (1 epoch) & 0.00 & 96.46 & 25.00 & 98.79 & 27.01 & 24.24 &4.95E4\\
  Linear (5 epochs) & 50.00 & 99.57 & \textbf{100.00} & \textbf{100.00} & 55.07 & 51.28 & 4.97E4\\
  
  Linear (converged) &\underline{75.00} & \underline{99.95} &\textbf{100.00}	& \textbf{100.00}	& \textbf{79.30}	& \textbf{68.60} & 5.33E4\\
  
  Fine-tune (1 epoch) & 25.00 & 99.05 & 25.00 & 99.47 & 19.61 & 15.52 &6.51E5\\
   Fine-tune (5 epochs) & 25.00 & 99.55 & {\bf 100.00} & {\bf 100.00} & 68.47 & 58.33 &4.28E6\\
\hline  
MI ($\alpha$=0.01)~\cite{poole2019variational} & \multirow{1}{*}{0.00} & \multirow{1}{*}{94.84} & \multirow{1}{*}{25.00} & \multirow{1}{*}{97.43} & \multirow{1}{*}{-29.41} & \multirow{1}{*}{-17.81} & \multirow{1}{*}{1.62E5}\\

MI ($\alpha$=0.50)  & \multirow{1}{*}{0.00} & \multirow{1}{*}{96.66} & \multirow{1}{*}{0.00} & \multirow{1}{*}{97.03} & \multirow{1}{*}{-11.36} &  \multirow{1}{*}{-10.21} & \multirow{1}{*}{1.62E5}\\

MI w/ PCA ($\alpha$=0.01) & \multirow{1}{*}{50.00} & \multirow{1}{*}{99.60} & \multirow{1}{*}{75.00} & \multirow{1}{*}{99.85} & \multirow{1}{*}{52.14} & \multirow{1}{*}{51.34} &  \multirow{1}{*}{5.58E4}\\

MI w/ PCA ($\alpha$=0.50)  & \multirow{1}{*}{0.00} & \multirow{1}{*}{96.68} & \multirow{1}{*}{50.00} & \multirow{1}{*}{99.52} & \multirow{1}{*}{23.73} & \multirow{1}{*}{17.09} & \multirow{1}{*}{5.58E4} \\  

\hline
LEEP~\cite{leep} & \underline{75.00} & 99.44 & \underline{75.00} & \underline{99.90} & 50.36 & 55.49 & \underline{378.31}\\

\hline
$\mathcal{N}$LEEP & \textbf{100.00} & \textbf{100.00} & \textbf{100.00} & \textbf{100.00} & \underline{72.84} & \underline{67.49} & \textbf{12.95}\\ 
\hline
\end{tabular}
\vspace{-7pt}
\end{table*}

\begin{table*}[h!]
\centering
\caption{Checkpoint ranking results on Group III, the checkpoints of heterogeneous architectures (GFLOPS excludes a forward pass on training data, which takes 2.73E5 GFLOPS shared by all)}
\label{tab:comparision_arch}

\begin{tabular}{l|c|c|c|c|c|c|c}
\hline
 Method & Recall@$1$  & Rel@$1$  & Recall@$3$ & Rel@$3$  & Pearson & Kendall & GFLOPS  \\


\hline
   Linear (1 epoch)& 25.00 & 98.17 & 25.00 & 99.35 & 30.14 & 13.80 &  3.37E4\\
   Linear (5 epoch)& 25.00 & 98.98 & 25.00 & 99.63 & 33.45 & 18.95 &  3.38E4\\

 Linear (converged) &25.00 & \textbf{99.66} &25.00	& 99.72	& \underline{63.55}	&36.91 & 3.62E4\\
  Fine-tune (1 epoch)&  0.00 & 98.28 & 25.00 & 99.80 & 17.61 & 11.59 &4.43E5 \\
  Fine-tune (5 epoch)& 25.00 & 98.62 & 25.00 & 99.68 & 25.72 & 15.72 & 2.91E6\\
\hline  
%


MI ($\alpha$=0.01)~\cite{poole2019variational} & 25.00 & 98.29 & 25.00 & 99.34 & 4.42 & 2.94 &1.30E5\\

MI ($\alpha$=0.50)  & \multirow{1}{*}{25.00} & \multirow{1}{*}{98.36} & \multirow{1}{*}{25.00} & \multirow{1}{*}{99.37} &\multirow{1}{*}{-9.79}  & \multirow{1}{*}{-6.81} &1.30E5\\

MI w/ PCA ($\alpha$=0.01) & \multirow{1}{*}{0.00} & \multirow{1}{*}{\underline{99.18}} & \multirow{1}{*}{\underline{50.00}} & \multirow{1}{*}{\underline{99.82}} & \multirow{1}{*}{61.94} & \multirow{1}{*}{{38.83}} &5.56E4\\

MI w/ PCA ($\alpha$=0.50) & \multirow{1}{*}{0.00} & \multirow{1}{*}{96.34} & \multirow{1}{*}{0.00} & \multirow{1}{*}{98.47} & \multirow{1}{*}{33.17} & \multirow{1}{*}{21.26} &5.56E4\\  

\hline
LEEP~\cite{leep} & 25.00 & 97.36 & \textbf{75.00} & \textbf{99.90} & 42.99 & \underline{45.06} & \underline{247.56}\\ 

\hline
$\mathcal{N}$LEEP & 25.00 &\textbf{99.66} & 25.00 & 99.70 & \textbf{66.94} & \textbf{51.14} & \textbf{12.68}\\ 

\hline
\end{tabular}
\vspace{-7pt}
\end{table*}

%% file: related.tex
\section{Related Work}
Our work is broadly related to task transferability and neural networks' generalization gap. 

{\noindent \bf Task transferability.}
A task usually refers to a joint distribution over input and label. Task transferability aims to predict how well a deep neural network pre-trained on a source task transfers to the target task. One may estimate the task transferability by data similarities regardless of models being used. Some work in this line includes conditional entropy~\cite{tran2019transferability}, data set distance as optimal transport~\cite{alvarez2020geometric},   $F$-relatedness~\cite{ben2003exploiting}, $A$-distance~\cite{kifer2004detecting}, and discrepancy distance~\cite{mansour2009domain}. Besides, Poole~\textit{et al.}~\cite{poole2019variational} derived information theoretic bounds. These methods are generally hard to compute in practice and rely on the availability of the source data. Some recent task transferability estimators involve both data and the models. Taskonomy~\cite{zamir2018taskonomy} is a fully computation method, where task similarity scores are obtained by transfer learning experiments.  Dwivedi~\textit{et al.}~\cite{dwivedi2019representation} analyzed the representation similarities to construct a task taxonomy. 
Besides the models trained on source tasks, all these methods also require a fine-tuned or independently trained model from the target task. In contrast, our work aims to find  checkpoint ranking measures that are lightweight in computing and requires no access to the source tasks. 

Recent works demonstrated that using pre-trained checkpoints that have similar feature representations as the target task's representations can improve transfer learning~\cite{dwivedi2019representation,song2019deep,song2020depara}. Song~\textit{et al.}~\cite{song2019deep,song2020depara} employed attribution maps to compare two models and then quantified transferabilities by the similarity of two models. 
Those approaches all require a converged model on target datasets, incurring intensive computation. 
However, we want to design a lightweight method for ranking checkpoints, ideally without any training procedures.

{\noindent \bf Predicting neural networks' generation gap.}
The difference between a model's performance on the training data versus its performance on test data is known as the generalization gap. It is practically useful and theoretically impactful to predict a neural network's generalization gap. Most recent work does so by finding a set of features that is predictive of the generalization, e.g.,  by estimating data margins~\cite{bartlett2017spectrally, elsayed2018large,sokolic2017robust}. Jiang~\textit{et al.}~\cite{jiang2018predicting} and Yak~\textit{et al.}~\cite{yak2019towards} demonstrate how the margin signatures of a neural network can predict the generalization gap with small errors. Besides, the network complexity and noise stability are also useful cues~\cite{neyshabur2017exploring, kawaguchi2017generalization, bartlett2017spectrally,arora2018stronger}. Our problem substantially differs from predicting the neural networks' generalization gap, which is concerned with the training and test data sets that share the same underlying distribution. We instead care about the results after fine-tuning a network's checkpoint.


%% file: supp_concat.tex
\section{Appendix}

In this appendix, we provide the following details to support the main text:
\begin{description}
\item[Section~\ref{app:tasks}:] Descriptions of the 4 downstream tasks.
\item[Section~\ref{app:sweep}:] Training details of pre-training and fine-tuning.
\item[Section~\ref{app:cmpall}:] Comparison results on the combined group of checkpoints in Groups I, II and III. 

\item[Section~\ref{app:cmp101s}:] Another group of checkpoints with ResNet101s at different pre-training stages.
\item[Section~\ref{app:moreon1-4}:] More experiment results on Groups I-IV.

\item[Section~\ref{app:decandseg}:] Neural checkpoints ranking on object detection and instance segmentation.
\end{description}

\subsection{Downstream tasks}\label{app:tasks}
In this section, we describe the datasets used for the downstream tasks as shown in Table~\ref{tab:tasks}. More specifically, \textbf{Caltech101}~\cite{fei2006one} contains 101 classes, including animals, airplanes, chairs and etc, the image size varies from 200 to 300 pixels per edge. \textbf{Flowers102}~\cite{nilsback2008automated} have 102 classes, with 40 to 248 training images per class, each image has at least 500 pixels. \textbf{Patch Camelyon}~\cite{veeling2018rotation} contains 327,680 images of histopathologic scans of
lymph node sections with image size of 96x96, which is collected to predict the presence of metastatic tissue. \textbf{Sun397}~\cite{xiao2010sun} is a scenery benchmark with 397 classes, including cathedral, staircase, shelter, river, or archipelago. There are at least 100 images per class. The images are in 200x200 or higher resolutions. We believe the dataset portfolio well represents a broad set of vision tasks.

\subsection{Hyper-parameter Sweep}\label{app:sweep}
We adopt the similar experiment setting as in~\cite{zhai2019visual} to fine-tune the neural networks on the downstream tasks. Specifically, we set the batch size to 512 and use SGD with momentum of 0.9. We do not use weight decay for fine-tuning, and we set it to be 0.01 times the learning rate~\cite{loshchilov2017decoupled} when training from scratch. We perform per-task hyper-parameter search. For each task, we sweep the learning rate in $\{$0.0001, 0.001, 0.005, 0.01, 0.05, 0.1, 0.2, 0.5$\}$ and the training step in $\{$2500, 5000, 10000, 15000, 20000, 400000$\}$. We incorporate inception data augmentation~\cite{szegedy2016inception} for pre-training checkpoints and we do not use  data-augmentation when we fine-tune 
the neural networks on the downstream tasks to emphasize the effect of transfer learning.

\subsection{Comparison results on all checkpoints in Groups I, II, III }\label{app:cmpall}

To obtain a comprehensive analysis, we also consolidate the checkpoints from Group I, II and III into one group (including 41 checkpoints in total) and then apply the ranking methods on it. Table~\ref{tab:comparision_all} shows the comparison results. The results further evaluate our observations in Section 4 of the main text. $\mathcal{N}$LEEP performs consistently well on the big group of checkpoints with lowest computation cost.  Linear separability of the feature representation is also a good indicator for ranking a large group of neural checkpoints. Fine-tuning with early stopping and mutual information estimator produce poor correlations. The ranking qualities of different ranking methods on the large group of checkpoints are in sharper contrast than on small groups. For instance, the Pearson's $r$ of $\mathcal{N}$LEEP \textit{vs}. Finetune (5 epochs) on the large group is 83.71 \textit{vs}. 27.84  but they perform 72.84 \textit{vs}. 68.47 on Group II (Table 2 in the main text). It indicates that $\mathcal{N}$LEEP is a low-variance and low-bias checkpoint ranking estimator, while early stopping may produce high-variance ranking results.

\subsection{Group IV: Supervised ResNet101s}\label{app:cmp101s}
We incorporate another group of checkpoints, including 12 ResNet101~\cite{he2016deep} models pre-trained by fully supervised learning on ImageNet~\cite{deng2009imagenet},  iNaturalist~\cite{van2018inaturalist}, and Places-365~\cite{zhou2017places}. We obtain the checkpoints in the same way as we have done for Group II, but with ResNet101 architecture. We want to study how different model architecture and model size affect the ranking quality.

\begin{figure*}[htbp]
\centering
\includegraphics[width = 0.9\textwidth]{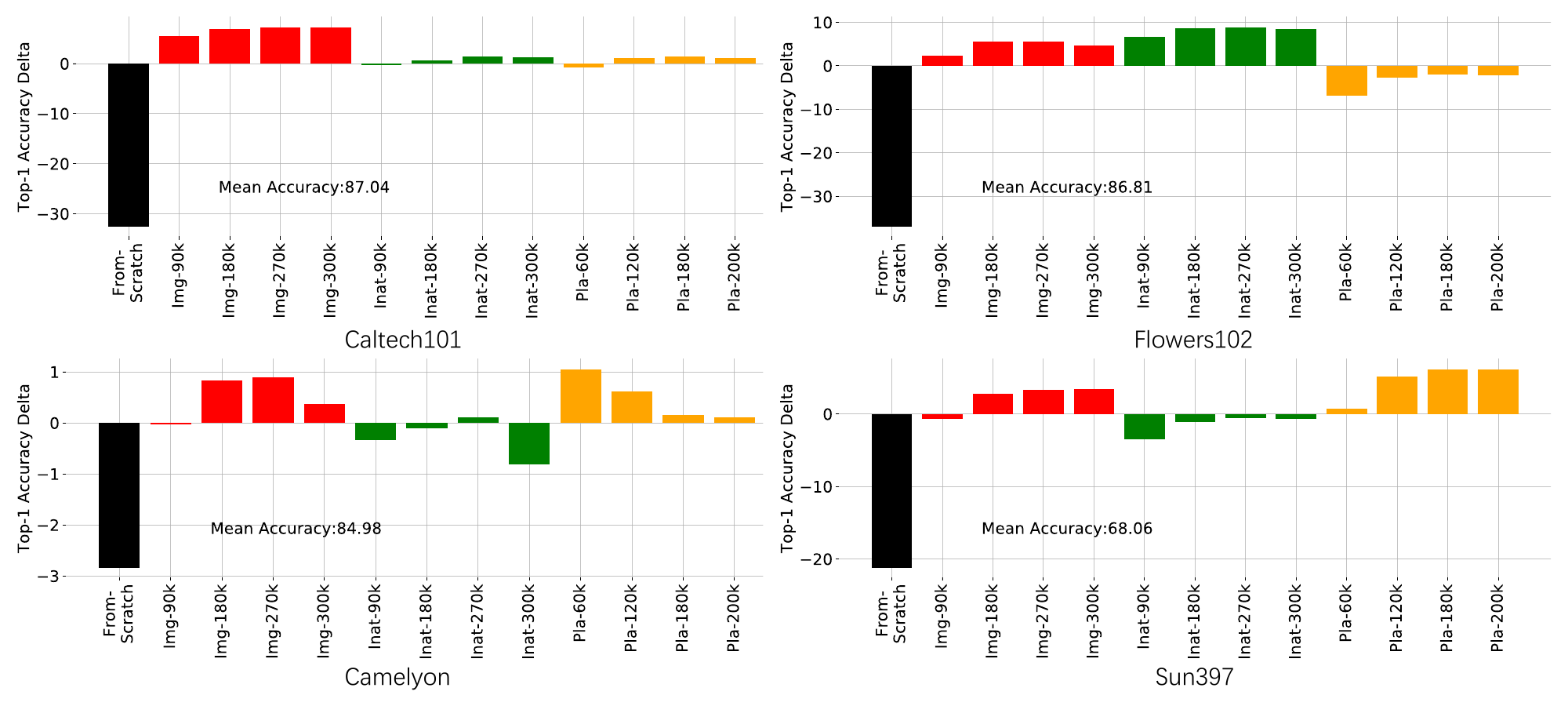}
\vspace{-10pt}
\caption{Difference between the fine-tuning accuracy of each checkpoint and the mean fine-tuning accuracy on Group IV. Black bar means From-Scratch. Red, green and orange bars represent ImageNet models, iNaturalist models and Places365 models, respectively. Img-90k means the checkpoint obtained by early stopping at the 90k-th iteration on ImageNet, and so on.}
\label{fig:groundtruth_sup_101}
\vspace{-3pt}
\end{figure*}

Figure~\ref{fig:groundtruth_sup_101} and Table~\ref{tab:absgroundtruth_sup_101} show the fine-tuning accuracy on 4 downstream tasks. The relative fine-tuning accuracies are similar to the accuracies on Group II. We also observe that a converged checkpoint does not necessarily demonstrates the best performance on the downstream tasks (cf. Img-270k is better than Img-300k on Flowers102~\cite{nilsback2008automated}).  Table~\ref{tab:comparision_sup101} shows the comparison results of ranking methods on those checkpoints. The relative performance among the ranking methods is similar to what they do in Group II (Table 2 in the main text). Except that they perform better on ResNet101s, e.g., Linear (converged) can achieve 68.60 in terms of Kendall's $\tau$ on ResNet50s versus 73.48 on ResNet101s, $\mathcal{N}$LEEP can get 72.84 in terms of Pearson's $r$ on ResNet50s versus 83.22 on ResNet101s. The observation reveals that the ranking of deeper checkpoints may be more predictable than shallow ones.

\subsection{More experimental results on Groups I-IV}\label{app:moreon1-4}
\begin{figure*}[h!]
\centering
\includegraphics[width = 1.0\textwidth]{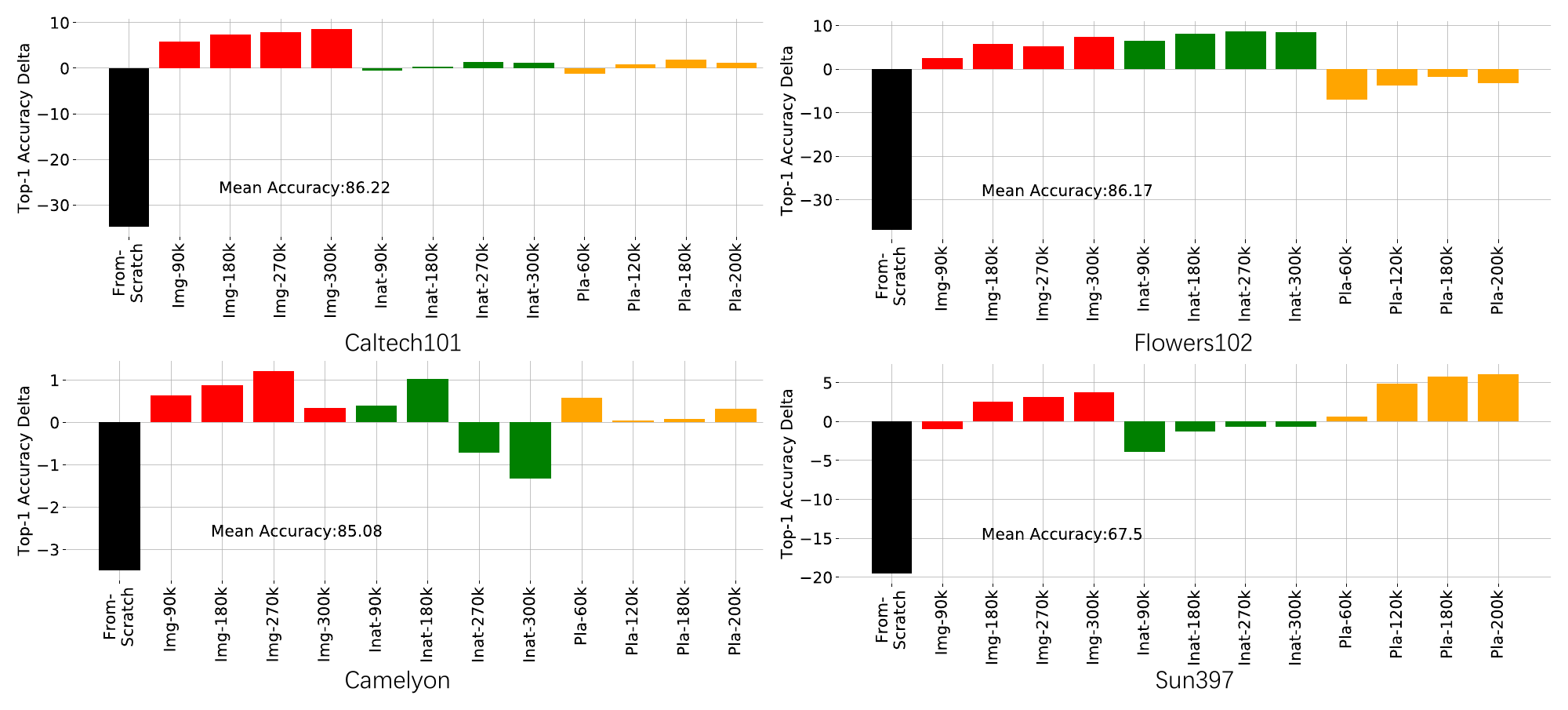}
\caption{Difference between the fine-tuning accuracy of each checkpoint and the mean fine-tuning accuracy on Group II. Black bar means From-Scratch. Red, green and orange bars represent ImageNet models, iNaturalist models and Places365 models, respectively. Img-90k means the checkpoint obtained by early stopping at the 90k-th iteration on ImageNet, and so on.}
\label{fig:groundtruth_sup}
\end{figure*}

\begin{figure*}[htbp]
\centering
\includegraphics[width = 1.0\textwidth]{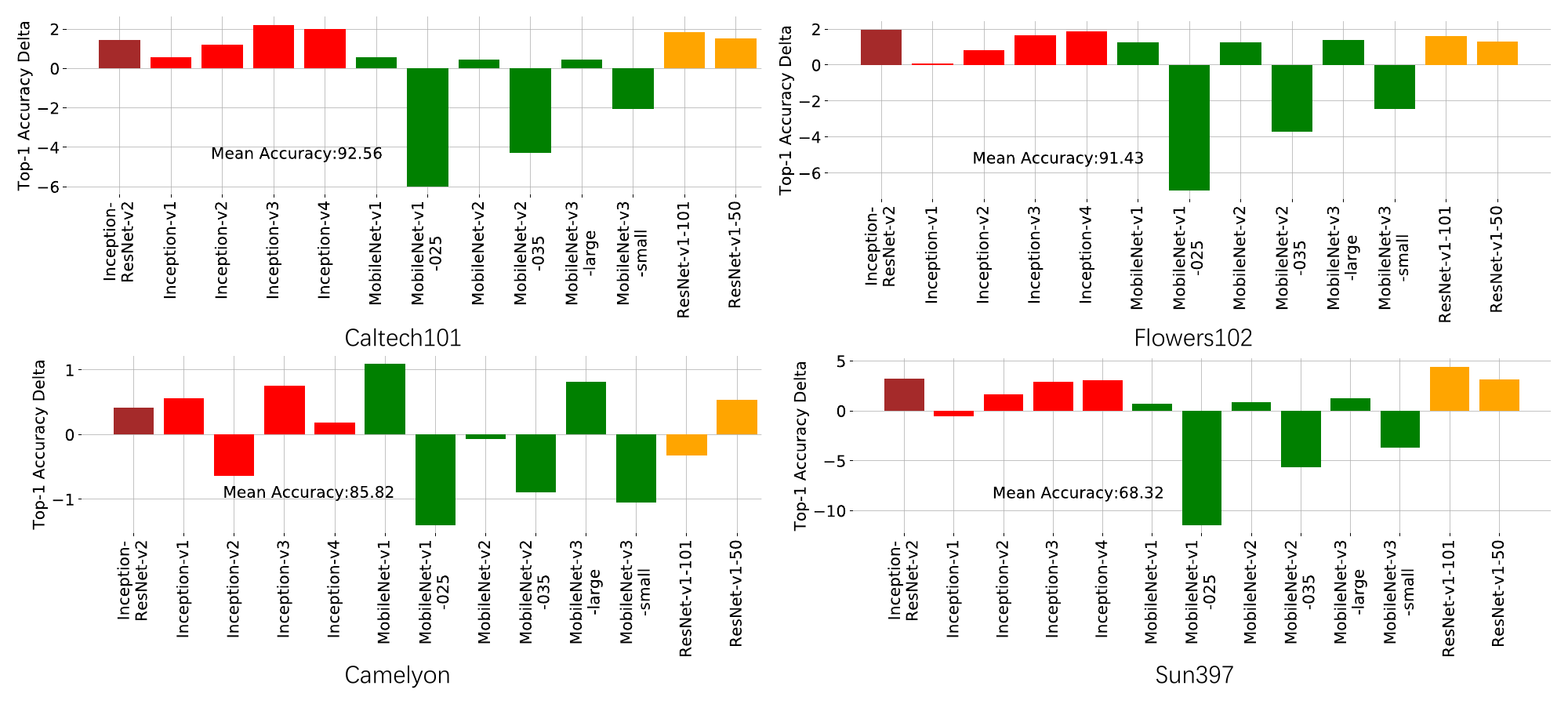}
\caption{Difference between the fine-tuning accuracy of each checkpoint and the mean fine-tuning accuracy on  Group III.  The colors of bars represent the models trained with different architectures. Brown: Inception-ResNet-V2. Red: Inception family. Green: MobileNet family and their variants. Orange: ResNet-v1 family.}
\label{fig:groundtruth_sup_arch}
\end{figure*}

We show more comparison results on \benchmark in this section. Figures~\ref{fig:groundtruth_sup} and~\ref{fig:groundtruth_sup_arch} show the best fine-tuning accuracies offset by their mean (for better visualization) on Groups II and III, respectively. Table~\ref{tab:absgroundtruth_mixed}, \ref{tab:absgroundtruth_sup}, \ref{tab:absgroundtruth_sup_arch}, \ref{tab:absgroundtruth_sup_101} demonstrate the absolute best fine-tuning accuracies on Groups I-IV, respectively.

\subsection{Neural checkpoint ranking for object detection and instance segmentation}\label{app:decandseg}

We also evaluate
on object detection and segmentation tasks, and show the results in Tables~\ref{tab:voc}. 
Specifically, we incorporate the recent self-supervised MoCo models (MoCov1, MoCov2, MoCov2-800epoch) and a ResNet50 model (supervised pretrained on ImageNet) into a new group of checkpoints. 
We evaluate checkpoint ranking on Pascal VOC (object detection) and Cityscapes (instance segmentation). 
In order to adapt $\mathcal{N}$LEEP to detection and segmentation tasks, we assign multiple ground truth labels for one image if it includes multiple object categories and extract the image-level features to perform GMM. 
We adapt $\mathcal{N}$LEEP to detection and segmentation tasks by assigning multi-labels to images with multiple object categories. The experiment results demonstrate that $\mathcal{N}$LEEP consistently outperforms the fine-tune and linear evaluation based approachs. We plan to include more diverse downstream tasks in \benchmark to facilitate future research.

\begin{table*}[htbp]
\centering
\vspace{-15pt}
\caption{Statistics of the datasets associated with the downstream tasks}
\label{tab:tasks}
\begin{tabular}{l|c|c|c}
\hline
 Dataset & Training  & Evaluation  & Number of Classes \\
 \hline
 Caltech101~\cite{fei2006one}	& 3060 &6084 & 101 \\
Flower102~\cite{nilsback2008automated} &	2040	&6149 & 102 \\
Patch-Camelyon~\cite{veeling2018rotation}	& 262144 &32768 & 2 \\
Sun397~\cite{xiao2010sun} &76128 &	10875 & 397 \\
  
\hline
 
\end{tabular}
\vspace{-8pt}
\end{table*}

\begin{table*}[htbp]
\centering
\vspace{-10pt}
\caption{Comparison results on all checkpoints in Group I, II, III (GFLOPS excludes a forward pass on training data, which takes 2.73E5 GFLOPS shared by all).}
\label{tab:comparision_all}
\begin{tabular}{l|c|c|c|c|c|c|c}
\hline
  Method & Recall@$1$  & Rel@$1$  & Recall@$3$ & Rel@$3$  & Pearson & Kendall & GFLOPS  \\


\hline
  Linear (1 epoch) &0.00 & 99.13 &\underline{25.00} &99.46 &22.30 &13.42 &\underline{4.45E4}\\
  Linear (5 epochs) &0.00 &99.13 &\underline{25.00} &99.21 &42.99 &31.64 & 4.47E4\\
  
  Linear (converged) &\underline{25.00} & 99.42 &\textbf{50.00}	& 99.73	& \underline{76.22}	& \underline{61.22} & 4.79E4\\
  
  Fine-tune (1 epoch) & 0.00 & 96.69 & 0.00 & 98.16 & 3.84 & 6.50 &5.85E5\\
   Fine-tune (5 epochs) & 0.00 & \textbf{99.49} & 0.00 & 99.49 & 27.20 & 27.16 &3.84E6\\

\hline  
MI ($\alpha$=0.01)~\cite{poole2019variational} & \multirow{1}{*}{0.00} & \multirow{1}{*}{77.50} & \multirow{1}{*}{0.00} & \multirow{1}{*}{81.44} & \multirow{1}{*}{1.12} & \multirow{1}{*}{7.16} & \multirow{1}{*}{1.52E5}\\

MI ($\alpha$=0.50)  & \multirow{1}{*}{0.00} & \multirow{1}{*}{66.51} & \multirow{1}{*}{0.00} & \multirow{1}{*}{90.07} & \multirow{1}{*}{-4.05} &  \multirow{1}{*}{-14.22} & \multirow{1}{*}{1.52E5}\\

MI w/ PCA ($\alpha$=0.01) & \multirow{1}{*}{0.00} & \multirow{1}{*}{89.18} & \multirow{1}{*}{\textbf{50.00}} & \multirow{1}{*}{\textbf{99.84}} & \multirow{1}{*}{12.14} & \multirow{1}{*}{20.99} &  \multirow{1}{*}{5.57E4}\\

MI w/ PCA ($\alpha$=0.50)  & \multirow{1}{*}{0.00} & \multirow{1}{*}{97.07} & \multirow{1}{*}{0.00} & \multirow{1}{*}{98.70} & \multirow{1}{*}{-14.03} & \multirow{1}{*}{-2.39} & \multirow{1}{*}{5.57E4} \\  

\hline
LEEP~\cite{leep} & \multirow{1}{*}{--} & \multirow{1}{*}{--} & \multirow{1}{*}{--} & \multirow{1}{*}{--} & \multirow{1}{*}{--} & \multirow{1}{*}{--} & \multirow{1}{*}{--}\\ 

\hline
$\mathcal{N}$LEEP & \textbf{50.00} & \underline{99.47} & \textbf{50.00} & \underline{99.78} & \textbf{83.71} & \textbf{68.18} & \textbf{12.86}\\ 
\hline
\end{tabular}
\vspace{-6pt}
\end{table*}

\begin{table*}[h]
\centering
\vspace{-8pt}
\caption{Comparison results on Group IV (GFLOPS excludes a forward pass on training data, which takes 6.27E5 GFLOPS shared by all).}
\label{tab:comparision_sup101}


\begin{tabular}{l|c|c|c|c|c|c|c}
\hline
  Method & Recall@$1$  & Rel@$1$  & Recall@$3$ & Rel@$3$  & Pearson & Kendall & GFLOPS  \\

\hline
  Linear (1 epoch) &0.00 & 98.58 &25.00 &98.99 &46.75 &27.27 &1.021E5\\
  Linear (5 epochs) &0.00 &98.72 &\underline{75.00} &\underline{99.95} &59.27 &41.32 & 1.023E5\\
  
  Linear (converged) &\underline{25.00} & 99.81 &\underline{75.00}	& \underline{99.95}	& \underline{82.17}	& \underline{73.48} & 1.06E5\\
  
  Fine-tune (1 epoch) & 0.00 & 96.19 & 25.00 & 99.34 & 29.64 & 21.21 &1.34E6\\
   Fine-tune (5 epochs) & \textbf{75.00} & \textbf{99.98} & \underline{75.00} & 99.94 & 69.19 & 50.00 &8.81E6\\

\hline  
MI ($\alpha$=0.01)~\cite{poole2019variational} & \multirow{1}{*}{0.00} & \multirow{1}{*}{97.25} & \multirow{1}{*}{\underline{75.00}} & \multirow{1}{*}{98.46} & \multirow{1}{*}{12.96} & \multirow{1}{*}{13.21} & \multirow{1}{*}{1.62E5}\\

MI ($\alpha$=0.50)  & \multirow{1}{*}{\underline{25.00}} & \multirow{1}{*}{98.60} & \multirow{1}{*}{50.00} & \multirow{1}{*}{99.54} & \multirow{1}{*}{30.16} &  \multirow{1}{*}{18.21} & \multirow{1}{*}{1.62E5}\\

MI w/ PCA ($\alpha$=0.01) & \multirow{1}{*}{0.00} & \multirow{1}{*}{\underline{99.85}} & \multirow{1}{*}{\underline{75.00}} & \multirow{1}{*}{\underline{99.95}} & \multirow{1}{*}{51.85} & \multirow{1}{*}{48.91} &  \multirow{1}{*}{5.58E4}\\

MI w/ PCA ($\alpha$=0.50)  & \multirow{1}{*}{0.00} & \multirow{1}{*}{95.99} & \multirow{1}{*}{50.00} & \multirow{1}{*}{98.41} & \multirow{1}{*}{48.64} & \multirow{1}{*}{44.31} & \multirow{1}{*}{5.58E4} \\  

\hline
LEEP~\cite{leep} & \multirow{1}{*}{\underline{25.00}} & \multirow{1}{*}{99.52} & \multirow{1}{*}{\underline{75.00}} & \multirow{1}{*}{99.72} & \multirow{1}{*}{54.54} & \multirow{1}{*}{46.43} & \multirow{1}{*}{\underline{378.31}}\\ 

\hline
$\mathcal{N}$LEEP & \textbf{75.00} & \textbf{99.98} & \textbf{100.00} & \textbf{100.00} & \textbf{83.22} & \textbf{73.80} & \textbf{12.95}\\ 
\hline
\end{tabular}
\vspace{-5pt}
\end{table*}

\begin{table*}[htbp]
\centering
\caption{Absolute fine-tuning accuracy on Group I.}
\label{tab:absgroundtruth_mixed}
\resizebox{\textwidth}{!}{%
\begin{tabular}{l|c|cccccc|cccc|cc|c|c|c|c}
\hline

\rotatebox{90}{Dataset} &\rotatebox{90}{From-Scratch} & \rotatebox{90}{WAE-UKL} & \rotatebox{90}{WAE-GAN} & \rotatebox{90}{Cond-BigGAN} &  \rotatebox{90}{WAE-MMD} & \rotatebox{90}{VAE} & \rotatebox{90}{Uncond-BigGAN} & \rotatebox{90}{Jigsaw} &\rotatebox{90}{Rel.Pat.Loc}  &\rotatebox{90}{Exemplar} &\rotatebox{90}{Rotation} & \rotatebox{90}{Semi-Rotation-10\%} & \rotatebox{90}{Semi-Exemplar-10\%} & \rotatebox{90}{Sup-100\%-Img} & \rotatebox{90}{Sup-Exemplar-100\%} & \rotatebox{90}{Sup-100\%-Inat} & \rotatebox{90}{Sup-100\%-Pla} \\
\hline
Caltech101 & 51.44 & 41.99 & 42.37 & 73.88 & 51.83 &54.91 & 73.15 &  78.85 & 79.09 & 80.02 &87.91 & 92.73& 92.77 & 94.42&  94.5&  87.00 &87.27 \\
Flowers102 & 49.28 & 17.46 &  17.96&  69.67& 26.05& 26.98& 61.72&  77.69& 78.08& 78.87& 83.29& 91.46& 91.28& 93.05 & 93.91& 94.47& 82.79 \\
Camelyon & 81.59& 80.55& 79.71& 80.73& 80.73& 80.87& 82.14&  85.43& 86.58& 85.27& 85.81& 85.09& 85.83& 85.37& 84.98& 83.33& 84.77 \\
Sun397 &47.97& 30.5 & 31.52 & 44.99& 37.56& 39.85& 47.36&  59.71& 58.06& 58.05& 60.00   & 66.74& 67.27& 70.98& 70.06& 66.65& 73.45 \\
\hline
\end{tabular}
}
\vspace{-8pt}
\end{table*}

\begin{table*}[htbp]
\centering
\caption{Absolute fine-tuning accuracy on Group II.}
\label{tab:absgroundtruth_sup}
\resizebox{\textwidth}{!}{%
\begin{tabular}{l|c|cccc|cccc|cccc}
\hline

\rotatebox{90}{Dataset} &\rotatebox{90}{From-Scratch} & \rotatebox{90}{Img-90k} & \rotatebox{90}{Img-180k} & \rotatebox{90}{Img-270k} &  \rotatebox{90}{Img-300k} & \rotatebox{90}{Inat-90k} & \rotatebox{90}{Inat-180k} & \rotatebox{90}{Inat-270k} &\rotatebox{90}{Inat-300k}  &\rotatebox{90}{Pla-60k} &\rotatebox{90}{Pla-120k} & \rotatebox{90}{Pla-180k} & \rotatebox{90}{Pla-200k} \\
\hline

Caltech101 & 51.44 &92.08 &93.55 &94.18 &94.73 &85.69 &86.54 &87.66 &87.43 &84.91 &87.11 &88.01 &87.48 \\

Flowers102 & 49.28 &88.8 &91.96 &91.36 &93.6 &92.71 &94.32 &94.78 &94.6 &79.15 &82.47 &84.36 &82.88 \\

Camelyon & 81.59 &85.73 &85.97 &86.3 &85.43 &85.48 &86.12 &84.37 &83.75 &85.67 &85.13 &85.16 &85.4 \\ 

Sun397 & 47.97 &66.56 &70.07 &70.69 &71.24 &63.63 &66.24 &66.84 &66.87 &68.14 &72.34 &73.25 &73.6 \\

\hline
\end{tabular}
}
\vspace{-8pt}
\end{table*}

\begin{table*}[htbp]
\centering

\caption{Absolute fine-tuning accuracy on Group III.}
\label{tab:absgroundtruth_sup_arch}
\resizebox{\textwidth}{!}{%
\begin{tabular}{l|c|cccc|cccccc|cc}
\hline
\rotatebox{90}{Dataset} &\rotatebox{90}{Inception-ResNet-v2} & \rotatebox{90}{Inception-v1} & \rotatebox{90}{Inception-v2} & \rotatebox{90}{Inception-v3} &  \rotatebox{90}{Inception-v4} & \rotatebox{90}{MobileNet-v1} & \rotatebox{90}{MobileNet-v1-025} & \rotatebox{90}{MobileNet-v2} &\rotatebox{90}{MobileNet-v2-035}  &\rotatebox{90}{MobileNet-v3-large} &\rotatebox{90}{MobileNet-v1-small} & \rotatebox{90}{ResNet-v1-101} & \rotatebox{90}{ResNet-v1-50} \\
\hline

Caltech101& 94.02 &93.13 &93.77 &94.76 &94.55 &93.15 &86.59 &93.02 &88.28 &93.02 &90.52 &94.42 &94.09 \\
Flowers102 & 93.39 &91.5 &92.25 &93.07 &93.31 &92.69 &84.44 &92.68 &87.71 &92.81 &89.01 &93.03 &92.72 \\
Camelyon & 86.23 &86.38 &85.18 &86.58 &86.0 &86.91 &84.42 &85.75 &84.93 &86.64 &84.77 &85.49 &86.35 \\
Sun397 & 71.52 &67.82 &69.95 &71.23 &71.41 &69.03 &56.88 &69.22 &62.7 &69.61 &64.63 &72.74 &71.44 \\

\hline
\end{tabular}
}
\vspace{-8pt}
\end{table*}

\begin{table*}[htbp]
\centering
\caption{Absolute fine-tuning accuracy on Group IV.}
\label{tab:absgroundtruth_sup_101}
\resizebox{\textwidth}{!}{%
\begin{tabular}{l|c|cccc|cccc|cccc}
\hline

\rotatebox{90}{Dataset} &\rotatebox{90}{From-Scratch} & \rotatebox{90}{Img-90k} & \rotatebox{90}{Img-180k} & \rotatebox{90}{Img-270k} &  \rotatebox{90}{Img-300k} & \rotatebox{90}{Inat-90k} & \rotatebox{90}{Inat-180k} & \rotatebox{90}{Inat-270k} &\rotatebox{90}{Inat-300k}  &\rotatebox{90}{Pla-60k} &\rotatebox{90}{Pla-120k} & \rotatebox{90}{Pla-180k} & \rotatebox{90}{Pla-200k} \\
\hline

Caltech101& 54.47 &92.56 &93.91 &94.23 &94.32 &86.68 &87.64 &88.51 &88.28 &86.22 &88.09 &88.44 &88.14\\

Flowers102 & 49.85 &89.13 &92.45 &92.37 &91.47 &93.46 &95.44 &95.7 &95.26 &79.96 &84.05 &84.72 &84.62\\

Camelyon & 82.14 &84.95 &85.81 &85.87 &85.35 &84.64 &84.87 &85.08 &84.16 &86.03 &85.59 &85.13 &85.09\\

Sun397 & 46.87 &67.36 &70.83 &71.41 &71.44 &64.6 &66.97 &67.44 &67.42 &68.78 &73.21 &74.22 &74.24 \\

\hline
\end{tabular}
}
\vspace{-8pt}
\end{table*}

\begin{table*}[htbp]
\begin{floatrow}
\caption{Left: Checkpoint ranking results on the Pascal VOC Object Detection Benchmark (trained on VOC 2007 train+val + VOC 2012 train+val, tested on VOC 2007 using AP). Right: Checkpoint ranking for Cityscapes instance segmentation.}
\label{tab:voc}

\begin{tabular}{l|c|c|c}
\hline
 Method  & Recall@1 & Pearson & Kendall   \\
 
  
\hline
  Linear (1 epoch)& \xmark & 18.45 &  -33.33 \\
  Linear (5 epoch)& \xmark & 40.77 & 0.00  \\

 Linear (converged)	& \cmark & 61.57 &  54.77 \\

  Fine-tune (1 epoch)& \xmark&20.55 & 0.00 \\
  Fine-tune (5 epoch)& \cmark& 50.46 & 33.33 \\
\hline  

\hline
$\mathcal{N}$LEEP & \cmark & \textbf{66.59} & \textbf{66.67}\\ 

\hline
\end{tabular}

\begin{tabular}{l|c|c|c}
\hline
 Method & Recall@1 & Pearson & Kendall   \\
 
  
\hline
   Linear (1 epoch)& \xmark &23.55& -33.33   \\
   Linear (5 epoch)& \xmark &55.43& 0.00  \\

 Linear (converged)	& \cmark &68.32 & 33.33 \\
 
  Fine-tune (1 epoch)& \xmark & 38.85 & -33.33\\
  Fine-tune (5 epoch)& \xmark & 51.22 & 33.33 \\
\hline  

\hline
$\mathcal{N}$LEEP & \cmark& \textbf{82.66} & \textbf{66.67}\\ 

\hline
\end{tabular}

\end{floatrow}
\end{table*}